\documentclass{cstr}

\usepackage{amsmath,amsfonts,amssymb,latexsym,bm}
\usepackage{amsthm}
\usepackage{mathrsfs}
\usepackage[dvipdfmx]{graphicx}
\usepackage{subfig}
\usepackage{url}
\usepackage{cite}

\usepackage{booktabs}
\usepackage{color}

\usepackage{algorithm, algorithmic}
\renewcommand{\algorithmicrequire}{{\bf Input (for user-side):}}
\renewcommand{\algorithmicensure}{{\bf Output (for user-side):}}

\title{
Interpretable collaborative data analysis on distributed data
}

\author[1,*]{Akira Imakura}
\author[1]{Hiroaki Inaba}
\author[1]{Yukihiko Okada}
\author[1]{Tetsuya Sakurai}

\affil[1]{University of Tsukuba, 1-1-1 Tennodai, Ibaraki, Tsukuba 305-8573, Japan}

\email{imakura@cs.tsukuba.ac.jp}

\begin{document}
\maketitle
\thispagestyle{titlepage}

\begin{abstract}
This paper proposes an interpretable non-model sharing collaborative data analysis method as one of the federated learning systems, which is an emerging technology to analyze distributed data.
Analyzing distributed data is essential in many applications such as medical, financial, and manufacturing data analyses due to privacy, and confidentiality concerns.
In addition, interpretability of the obtained model has an important role for practical applications of the federated learning systems. 
By centralizing {\it intermediate representations}, which are individually constructed in each party, the proposed method obtains an interpretable model, achieving a collaborative analysis without revealing the individual data and learning model distributed over local parties.
Numerical experiments indicate that the proposed method achieves better recognition performance for artificial and real-world problems than individual analysis.
\end{abstract}

\section{Introduction}
\subsection{Motivation}
In many applications, e.g., medical, financial, and manufacturing data analyses, sharing the original data for analysis may be difficult due to privacy and confidentiality requirements.
Distributed data analyses without revealing the individual data have recently attracted significant attention resulting in the federated learning systems including model share-type federated learning \cite{li2019survey,konevcny2016federated2,konevcny2016federated,mcmahan2016communication} and non-model share-type collaborative data analysis \cite{imakura2020data, ye2019distributed,imakura2020collaborative,bogdanova2020collaborative}.
In addition, for practical applications, it is known that the interpretability (i.e., ``the degree to which a human can understand the cause of a decision'' according to Miller's definition \cite{miller2017explanation}) of the obtained model plays an important role \cite{molnar2019interpretable, guidotti2018survey}.
\par
A motivating example would be the distributed medical data analysis for employees of companies.
In this scenario, employees (i.e., data samples) are distributed in multiple companies.
Their medical and work records (i.e., features) are distributed in multiple parties, e.g., the records of medical treatment and check are distributed in different medical institutions and the work situations of the employees are stored in the company's personnel department.
Due to the limited number of samples and features, the data in one party of one company could lack some useful information for analysis.
Centralizing the data from multiple parties for collaborative analysis could help to learn more useful information and obtain high-quality predictions.
However, due to privacy concerns, it is difficult to share individual medical records and work situations from multiple parties.
A similar situation occurs in financial and manufacturing data analyses.
Thus, collaborative data analysis for distributed data, which are partitioned according to samples and features, is essential and important.
\par
Moreover, when companies aim to adopt policies or decisions according to analyses of machine-learning systems, the model should be interpretable; i.e., people need to understand the reasons why the system obtained such results \cite{barrieta2020explainable}.
This will allow people to make more useful decisions.
Therefore, when distributed data analysis is used as a tool to support decision making, the model needs to be interpretable.
\subsection{Main purpose and contributions}
Federated learning is based on deep neural networks and data collaboration analysis constructs the multi-layer model via intermediate representations.
Thus, the interpretability of the obtained model is not high, which could limit its use in some application areas.
To the best of our knowledge, there have been limited investigations on interpretable model construction for distributed data in the literature.
\par
To meet the above needs of distributed data analysis and interpretability, we propose an interpretable non-model sharing collaborative data analysis on distributed data.
The proposed method generates dimensionally-reduced {\it intermediate representations} from individual data in local parties, which are then shared instead of the individual data and models.
The proposed method constructs an interpretable model for each party.
\par
The main contributions of this paper are summarized as follows:
\begin{itemize}
        \item The proposed method generates an interpretable model for distributed data based on sharing intermediate representations without revealing the private data and sharing the model.
        \item The obtained interpretable model is based on the whole features of distributed data, which is not possible to do in individual analysis.
        \item Each party can individually select an interpretable model according to its own needs.
        \item Numerical experiments on both artificial and real-world data show that the proposed method constructs an interpretable model with better recognition performance than individual analysis and comparable to that of centralized analysis.
\end{itemize}
\subsection{Outline of the paper}
In Section~\ref{sec:problem}, we state the target distributed data and review related works.
In Section~\ref{sec:propose}, we propose a novel interpretable collaborative data analysis.
Numerical results are reported in Section~\ref{sec:experiment}.
Finally, in Section~\ref{sec:conclusion}, we summarize the results and conclude the paper.
\par
Note that throughout the paper, we use the MATLAB colon notation to refer to ranges of matrix elements.
\section{Problem setting and related works}
\label{sec:problem}
\subsection{Problem setting}
In this paper, we consider the simple horizontal and vertical partitions.
However, we note that the proposed method can also applied to more complicated situations described in \cite{imakura2020collaborative}.
\par
Let $m$ and $n$ denote the numbers of features and training data samples.
In addition, let $X = [{\bm x}_{1}, {\bm x}_{2}, \dots, {\bm x}_{n}]^{\rm T} \in \mathbb{R}^{n \times m}$ and $Y = [{\bm y}_1, {\bm y}_2, \dots, {\bm y}_n]^{\rm T} \in \mathbb{R}^{n \times \ell}$ be the training dataset and the corresponding ground truth, respectively.
The $n$ data samples are partitioned into $c$ institutions and the $m$ features are partitioned into $d$ parties as follows:
\begin{equation}
  X = \left[
    \begin{array}{cccc}
      X_{1,1} & X_{1,2} & \cdots & X_{1,d} \\
      X_{2,1} & X_{2,2} & \cdots & X_{2,d} \\
      \vdots & \vdots & \ddots & \vdots \\
      X_{c,1} & X_{c,2} & \cdots & X_{c,d}
    \end{array}
  \right], \quad
  Y = \left[
    \begin{array}{c}
      Y_{1} \\
      Y_{2} \\
      \vdots \\
      Y_{c} 
    \end{array}
  \right].
\end{equation}
Then, the $(i,j)$-th party has partial dataset and the corresponding ground truth,
\begin{equation}
  X_{i,j} \in \mathbb{R}^{n_i \times m_j}, \quad Y_i \in \mathbb{R}^{n_i \times \ell}.
\end{equation}
\par
{\it Individual analysis} of the dataset in a local party may not have high-quality predictions due to the lack of feature information or insufficient samples.
If the datasets can be centralized from multiple parties and analyze them as one dataset, i.e., {\it centralized analysis}, then we expect to achieve high-quality predictions.
However, it is difficult to share individual data for centralization due to privacy and confidentiality concerns.
\par
All parties want to obtain an interpretable model to achieve the competitive prediction results as centralized analysis without sharing the private dataset $X_{i,j}$.
\subsection{Distributed data analysis}
Typical techniques for privacy-preserving distributed data analysis include cryptographic computations (or secure multi-party computation) \cite{jha2005privacy,cho2018secure,gilad2016cryptonets} e.g., using fully homomorphic encryption \cite{gentry2009fully}, and methods using differential privacy \cite{abadi2016deep,ji2014differential,dwork2006differential}, where randomization is used to protect the privacy of the original datasets.
\par
Recently, federated learning has been actively studied for distributed data analysis \cite{konevcny2016federated,konevcny2016federated2,mcmahan2016communication,yang2019federated}, where the learning model is centralized while the original datasets remain distributed in local parties.
Google first proposed the concept of federated learning in \cite{konevcny2016federated,konevcny2016federated2}, which is typically used for Android phone model updates \cite{mcmahan2016communication}.
Recently, there have been several efforts to improve federated learning, e.g., see \cite{li2019survey,yang2019federated} and reference therein.
Note that, for federated learning, we may need to care a privacy of the original dataset due to the shared functional model \cite{yang2019gdpr}.
Hence, non-model sharing-type method i.e., {\it collaborative data analysis}, have been proposed for supervised learning \cite{imakura2020data,imakura2020collaborative} and feature selection \cite{ye2019distributed}.
The performance comparison between collaborative data analysis and federated learning are reported in \cite{bogdanova2020collaborative}.
\subsection{Needs for interpretability in machine learning}
In recent years, as machine learning has been used in various application in society, there has been an active discussion to develop interpretable machine learning \cite{barrieta2020explainable}.
While regressions, rules, and decision trees have been considered to be interpretable in machine-learning models, decision trees in particular have long been used in the context of decision support due to the high transparency \cite{barrieta2020explainable,guidotti2018survey}.
Also, the need for model transparency from various stakeholders has increased to replace  high-performance black-box models currently used for making predictions \cite{preece2018stakeholders}.
Hence, to create an interpretable model with high prediction accuracy, researchers have developed interpretable models that mimic the behavior of black-box models.
\par
In response to these current needs, there is an opinion that since interpretability is a domain-specific concept, it is necessary to build models that consider ease of use and data structure \cite{rudin2019stop}.
Therefore, interpretable model construction algorithms needs to be designed such that users allowed to freely set the model according to its own needs.
\section{Interpretable collaborative data analysis}
\label{sec:propose}
Here, we briefly introduce the algorithm of collaborative data analysis \cite{imakura2020collaborative} and propose a novel interpretable non-model sharing collaborative data analysis on distributed data.
\subsection{Collaborative data analysis}
Collaborative data analysis has been proposed in \cite{imakura2020data,imakura2020collaborative} for distributed data together with a practical operation strategy to address privacy and confidentiality concerns.
Here, we briefly introduce the algorithm based on the practical operation strategy.
\par
In the practical operation strategy, collaborative data analysis is operated by two roles: {\it user} and {\it analyst}.
Users have the private dataset $X_{i,j}$ and the corresponding ground truth $Y_i$, which need to be analyzed without sharing $X_{i,j}$.
Each user individually constructs an dimensionally-reduced intermediate representation and shares it to the analyst.
To allow each user to use an individual function for generating intermediate representation, the analyst transforms the shared intermediate representations to an incorporable form called {\it collaboration representations} and analyzes them as one dataset.
\subsubsection{Training Phase: Construction of intermediate representations}
Each user constructs the intermediate representation,
\begin{align*}
  \widetilde{X}_{i,j} = f_{i,j}(X_{i,j}) \in \mathbb{R}^{n_i \times \widetilde{m}_{i,j}},
\end{align*}
where $f_{i,j}$ denotes a linear or nonlinear row-wise mapping function.
A typical setting for $f_{i,j}$ is dimensionality reduction, with $\widetilde{m}_{i,j} < m_{i,j}$, including unsupervised \cite{pearson1901liii, he2004locality, maaten2008visualizing} and supervised methods \cite{fisher1936use, sugiyama2007dimensionality, li2017locality, imakura2019complex}.
To address privacy and confidentiality concerns, the function $f_{i,j}$ should be set as
\begin{itemize}
  \item The private data $X_{i,j}$ can be obtained only if anyone has both the corresponding intermediate representation $\widetilde{X}_{i,j}$ and the mapping function $f_{i,j}$ or its approximation.
  \item The mapping function $f_{i,j}$ can be approximated only if anyone has both the input and output of $f_{i,j}$.
\end{itemize}
\par
Then, the resulting intermediate representations $\widetilde{X}_{i,j}$ are centralized to the analyst instead of the original private data $X_{i,j}$ or the trained model.
By sharing the intermediate representations $\widetilde{X}_{i,j}$ while keeping the mapping function $f_{i,j}$ in each party, the collaborative data analysis can address the privacy and confidentiality concerns.
\subsubsection{Training Phase: Construction and analysis of collaboration representations}
\label{sec:CDA}
Since $f_{i,j}$ depends on the user $(i,j)$, the analyst cannot analyze the shared intermediate representations as one dataset.
To overcome this problem, the intermediate representations $\widetilde{X}_{i,j}$ are transformed to incorporable collaboration representation as follows:
\begin{align*}
  \widehat{X}_{i} = g_{i}(\widetilde{X}_i) \in \mathbb{R}^{n_i \times \widehat{m}}, \quad
  \widetilde{X}_i = [\widetilde{X}_{i,1}, \widetilde{X}_{i,2}, \dots, \widetilde{X}_{i,d}] \in \mathbb{R}^{n_i \times \widetilde{m}_i},
\end{align*}
where a row-wise mapping function $g_i$ with $\widetilde{m}_i = \sum_{j=1}^d \widetilde{m}_{i,j}$ and $\widehat{m} = \min_i \widetilde{m}_{i}$.
\par
To construct the mapping function $g_i$ for incorporable collaboration representations, an {\it anchor dataset} $X^{\rm anc} \in \mathbb{R}^{r \times m}$, which is a shareable dataset consisting of public data or dummy data randomly constructed, is introduced.
The anchor dataset is shared to all users and is partitioned according to features, i.e.,
\begin{equation*}
  X^{\rm anc} = [X^{\rm anc}_{:,1}, X^{\rm anc}_{:,2}, \dots, X^{\rm anc}_{:,d}],
\end{equation*}
where $X_{:,j}^{\rm anc} \in \mathbb{R}^{r \times m_j}$.
At the user-side, applying each mapping function $f_{i,j}$ to the corresponding subset of the anchor dataset, $X_{:,j}^{\rm anc}$ becomes
\begin{align*}
  \widetilde{X}_{i,j}^{\rm anc} = f_{i,j}(X_{:,j}^{\rm anc}) \in \mathbb{R}^{r \times \widetilde{m}_{i,j}},
\end{align*}
which is centralized to the analyst.
Then, the mapping function $g_i$ is constructed such that
\begin{equation*}
  \widehat{X}_i^{\rm anc} = g_i(\widetilde{X}_i^{\rm anc}) \in \mathbb{R}^{r \times \widehat{m}}
        \quad \mbox{s.t. }
        \widehat{X}_{i}^{\rm anc} \approx \widehat{X}_{i'}^{\rm anc} \quad
        (i \neq i'),
\end{equation*}
where $\widetilde{X}_i^{\rm anc} = [\widetilde{X}_{i,1}^{\rm anc}, \widetilde{X}_{i,2}^{\rm anc}, \dots, \widetilde{X}_{i,d}^{\rm anc}]$.
For computing $g_i$, authors of \cite{imakura2020data, imakura2020collaborative} introduced a practical method via a total least squares problem \cite{ito2016algorithm} when $g_i$ is linear and also indicated an idea when $g_i$ is nonlinear.
\par
Finally, the obtained collaboration representations $\widehat{X}_i$ can be analyzed as one dataset,
\begin{equation*}
  \widehat{X} = \left[
    \begin{array}{c}
      \widehat{X}_1 \\
      \widehat{X}_2\\
      \vdots \\
      \widehat{X}_{c}
    \end{array}
  \right] \in \mathbb{R}^{n \times \widehat{m}},
\end{equation*}
together with the shared ground truth $Y_i$ using supervised machine learning and deep learning methods.
This will obtain a model
\begin{equation*}
  h(\widehat{X}) \approx Y.
\end{equation*}
\subsubsection{Prediction Phase}
\label{sec:prediction}
Let $X^{\rm test} \in \mathbb{R}^{s \times m}$ be a test dataset partitioned according to features and samples as
\begin{equation*}
  X^{\rm test} = \left[
    \begin{array}{cccc}
      X^{\rm test}_{1,1} & X^{\rm test}_{1,2} & \cdots & X^{\rm test}_{1,d} \\
      X^{\rm test}_{2,1} & X^{\rm test}_{2,2} & \cdots & X^{\rm test}_{2,d} \\
      \vdots & \vdots & \ddots & \vdots \\
      X^{\rm test}_{c,1} & X^{\rm test}_{c,2} & \cdots & X^{\rm test}_{c,d}
    \end{array}
  \right],
\end{equation*}
where $X^{\rm test}_{i,j} \in \mathbb{R}^{s_i \times m_j}$.
Then, the intermediate representations $\widetilde{X}_{i,j}^{\rm test} = f_{i,j}(X_{i,j}^{\rm test})$ are constructed at the user-side, and are shared to the analyst.
In the analyst-side, the predictions $Y_i^{\rm test}$ of $X_i^{\rm test} = [X_{i,1}^{\rm test}, X_{i,2}^{\rm test}, \dots, X_{i,d}^{\rm test}]$ are obtained by
\begin{equation*}
Y_i^{\rm test} = h( g_i([\widetilde{X}_{i,1}^{\rm test}, \widetilde{X}_{i,2}^{\rm test}, \dots, \widetilde{X}_{i,d}^{\rm test}]))
\end{equation*}
via the intermediate and collaboration representations and are returned to the corresponding users.
\subsection{Derivation of an interpretable collaborative data analysis}
As shown in Section~\ref{sec:prediction}, the obtained model of the $i$-th institution is
\begin{equation*}
  Y_i^{\rm test} = h( g_i([f_{i,1}(X_{i,1}^{\rm test}), f_{i,2}(X_{i,2}^{\rm test}), \dots, f_{i,d}(X_{i,d}^{\rm test})])),
\end{equation*}
which is a multi-layer model via intermediate and collaboration representations.
The model is separately hold by the users and the analyst such that $f_{i,j}$ is only at the user-side, and $g_i$ and $h$ are only at the analyst-side.
Therefore, the interpretability of the model is not so high even though a highly interpretable model is used, e.g., the decision tree for $h$.
To address this, we propose an interpretable collaborative data analysis.
\par
We first revisit the anchor data $X^{\rm anc}$.
In collaborative data analysis, the anchor data are shareable data consisting of public data or dummy data randomly constructed, and are used for constructing collaborative representation (see Section~\ref{sec:CDA}).
\par
The basic concept of the proposed method is to mimic the multi-layer model of collaborative data analysis, that is,
\begin{enumerate}
  \item Predict the anchor data $X^{\rm anc}$ using collaborative data analysis.
  \item Construct an interpretable model with the anchor data $X^{\rm anc}$ and their predictions.
\end{enumerate}
\par
The predictions of the anchor data $X^{\rm anc}$ are
\begin{equation*}
  Y_i^{\rm anc} = h( g_i([f_{i,1}(X_{:,1}^{\rm anc}), f_{i,2}(X_{:,2}^{\rm anc}), \dots, f_{i,d}(X_{:,d}^{\rm anc})]))
\end{equation*}
for each $i$.
In the collaborative data analysis, the analyst holds $\widetilde{X}_{i,j}^{\rm anc} = f_{i,j}(X^{\rm anc}_{:,j})$.
Therefore, $Y_i^{\rm anc}$ can be obtained by $Y_i^{\rm anc} = h(g_i([\widetilde{X}_{i,1}^{\rm anc}, \widetilde{X}_{i,2}^{\rm anc},$ $\dots, \widetilde{X}_{i,d}^{\rm anc}])$ without the need for additional communication from users.
Note that additional communication may increase a privacy and confidentiality risks.
Regarding higher recognition performance, we can use another dataset for this purpose different from the anchor data for constructing $g_i$ in the collaborative data analysis.
Then, the predictions $Y_i^{\rm anc}$ of the anchor data $X^{\rm anc}$ are returned to the $i$-th user.
\par
At the user-side, an interpretable model is individually constructed as
\begin{equation*}
  Y_i^{\rm anc} \approx t_i(X^{\rm anc}),
\end{equation*}
where the obtained model $t_i$ depends on $i$.
Note that, since the anchor data have whole features of $X$ instead of the private dataset $X_{i,j}$, the obtained interpretable model $t_i$ is based on whole features.
For example, in the decision tree, $t_i$ can be whole features represented by branch, which are not feasible to do in an individual analysis that only uses $X_{i,j}$.
Here, each party can individually select an interpretable model according to its own needs.
This is an advantage of the proposed method for practical applications.
\par
Note that the performance of the proposed method depends on the choice of the anchor data $X^{\rm anc}$.
The simplest way to set $X^{\rm anc}$ is via a random matrix \cite{imakura2020data, imakura2020collaborative,bogdanova2020collaborative}.
However, to improve the performance, the anchor data need to preserve some statistics of $X$.
One practical idea is to generate $X_{i,j}^{\rm anc}$ for each private data by using methods such as the generative adversarial nets (GAN) \cite{goodfellow2014generative} and autoencoder based on (deep) neural network or dimensionality reduction with data augmentation.
Then, $X_{i,j}^{\rm anc}$ is shared with all users and $X^{\rm anc}$ is set as
\begin{equation*}
  X^{\rm anc} = [X^{\rm anc}_{:,1}, X^{\rm anc}_{:,2}, \dots, X^{\rm anc}_{:,d}] = \left[
    \begin{array}{cccc}
      X^{\rm anc}_{1,1} & X^{\rm anc}_{1,2} & \cdots & X^{\rm anc}_{1,d} \\
      X^{\rm anc}_{2,1} & X^{\rm anc}_{2,2} & \cdots & X^{\rm anc}_{2,d} \\
      \vdots & \vdots & \ddots & \vdots \\
      X^{\rm anc}_{c,1} & X^{\rm anc}_{c,2} & \cdots & X^{\rm anc}_{c,d}
    \end{array}
  \right].
\end{equation*}
We will investigate practical techniques for constructing a suitable anchor data in the future.
\par
The pseudo-code of the proposed method is summarized in Algorithm~\ref{alg:proposed}.
As shown in Algorithm~\ref{alg:proposed}, the proposed interpretable collaborative data analysis is based on the {\it one-path} algorithm, which does not require iteration steps with data communication.
\begin{algorithm}[!t]
\footnotesize
\caption{Interpretable collaborative data analysis}
\label{alg:proposed}
\begin{algorithmic}
  \REQUIRE $X_{i,j} \in \mathbb{R}^{n_i \times m_{j}}, Y_i \in \mathbb{R}^{n_i \times \ell}$, individually
  \ENSURE Interpretable models $t_i$ $(i = 1, 2, \dots, c)$, which depend on $i$
  \STATE
  \STATE
  \begin{tabular}{llcl}
    & \multicolumn{1}{c}{ {\it user-side} $(i,j)$} && \multicolumn{1}{c}{ {\it analyst-side} } \\ \cmidrule{2-2} \cmidrule{4-4}
    1:  & Generate $X^{\rm anc}_{i,j}$ and share to all users && \\
    2:  & Set $X^{\rm anc}$ and $X_{:,j}^{\rm anc}$ && \\
    3:  & Generate $f_{i,j}$ && \\
    4:  & Compute $\widetilde{X}_{i,j} = f_{i,j}(X_{i,j})$ && \\
    5:  & Compute $\widetilde{X}^{\rm anc}_{i,j} = f_{i,j}(X_{:,j}^{\rm anc})$ && \\
    6:  & Share $\widetilde{X}_{i,j}, \widetilde{X}_{i,j}^{\rm anc}$ and $Y_i$ to analyst & $\rightarrow$ & Get $\widetilde{X}_{i,j}, \widetilde{X}_{i,j}^{\rm anc}$ and $Y_i$ for all $i$ and $j$ \\
    7:  & && Set $\widetilde{X}_i$ and $\widetilde{X}_i^{\rm anc}$ \\
    8:  & && Construct $g_i$ from $\widetilde{X}_{i}^{\rm anc}$ for all $i$ \\
    9:  & && Compute $\widehat{X}_{i} = g_i(\widetilde{X}_{i})$ for all $i$ \\
    10: & && Set $\widehat{X}$ and $Y$ \\
    11: & && Analyze $\widehat{X}$ and get $h$ as $Y \approx h(\widehat{X})$ \\
    12: & && Compute $\widehat{X}_{i}^{\rm anc} = g_i(\widetilde{X}_{i}^{\rm anc})$ for all $i$ \\
    13: & && Compute $Y_i^{\rm anc} = h(\widehat{X}_{i}^{\rm anc})$ for all $i$ \\
    14: & Get $Y_i^{\rm anc}$ &$\leftarrow$& Return $Y_i^{\rm anc}$ to users \\
    15: & Analyze $X^{\rm anc}$ and get $t_i$ && \\
        & as $Y_i^{\rm anc} \approx t_i(X^{\rm anc})$ && \\
  \end{tabular}
\end{algorithmic}
\end{algorithm}
\subsection{Discussion on privacy and confidentiality}
In the proposed method (Algorithm~\ref{alg:proposed}), each user shares the local anchor data $X_{i,j}^{\rm anc}$ to other users and shares intermediate representations $\widetilde{X}_{i,j}, \widetilde{X}^{\rm anc}_{i,j}$ to the analyst.
We discuss how the privacy of the private data $X_{i,j}$ is preserved for both the users and the analyst.
Here, we assume that the users do not trust each other and want to protect their training data $X_{i,j}$ against honest-but-curious users and the analyst.
Hence, the users and the analyst will strictly follow the strategy, but they will try to infer as much information as possible.
We also assume that the analyst does not collude with any users.
\par
To ensure the privacy of $X_{i,j}$ against other users, each user shares the local anchor data $X_{i,j}^{\rm anc}$ to other users.
The local anchor data do not contain $X_{i,j}$ but may preserve some useful information.
The local anchor data are constructed by the users themselves using methods such as GAN and autoencoder with data augmentation.
Therefore, users can control the information although it may result a trade-off in the performance.
Note that collaborative data analysis works well even when using random anchor data, as demonstrated in \cite{imakura2020data,imakura2020collaborative,bogdanova2020collaborative}.
\par
To ensure the privacy of $X_{i,j}$ against the analyst, each user shares the intermediate representations $\widetilde{X}_{i,j}, \widetilde{X}_{i,j}^{\rm anc}$ to the analyst.
If analyst has the map function $f_{i,j}$ or its approximation, he/she can obtain an approximation of $X_{i,j}$.
However, the function $f_{i,j}$ is private and cannot be approximated by others because no one has both the input and output the of $f_{i,j}$.
Therefore, the analyst cannot obtain an approximation of $X_{i,j}$ from the intermediate representations.
\par
In our future studies, we will further analyze more details of the privacy of the proposed method.
\section{Experiments}
\label{sec:experiment}
This section evaluates the performance of the proposed interpretable collaborative data analysis (Algorithm~\ref{alg:proposed}) and compares it with those of interpretable centralized and individual analyses for classification problems.
Note that centralized analysis is considered as an ideal case since the private datasets $X_{i,j}$ cannot be shared in our target situation.
The proposed collaborative data analysis aims to achieve a better performance than individual analysis.
\par
We use a simple decision tree for the interpretable model.
In the proposed method, each intermediate representation is designed from $X_{i,j}$ using locality preserving projections (LPP) \cite{he2004locality} which is an unsupervised dimensionality reduction method.
We use a kernel version of ridge regression (K-RR) \cite{saunders1998ridge} with a Gaussian kernel for data analysis for collaborative analysis.
We set the regularization parameter of K-RR to $\lambda = 0.01$.
The local anchor data $X_{i,j}^{\rm anc}$ are constructed by a low-rank approximation based on singular value decomposition (SVD) with random perturbation and data augmentation.
We set $r= 2,500$ as the number of anchor data.
\par
We set the ground truth $Y$ as a binary matrix whose $(i,j)$ entry is 1 if the training data ${\bm x}_i$ are in class $j$ and $0$ otherwise.
This type of ground truth $Y$ has been applied to various classification algorithms, including ridge regression and deep neural networks \cite{bishop2006pattern}.
\par
In this paper, we evaluate the performance of methods in terms of normalized mutual information (NMI) \cite{strehl2002cluster} and accuracy (ACC).
Moreover, to evaluate the similarity of prediction model by individual and collaborative data analyses with those of centralized analysis, we use fidelity to centralized analysis under NMI (Fidelity to CA), that is 
\begin{equation*}
  {\rm NMI}(Y^{\rm IA}, Y^{\rm CA}), \quad
  {\rm NMI}(Y^{\rm CDA}, Y^{\rm CA}),
\end{equation*}
where the function ${\rm NMI}$ denotes the value of NMI between two predictions and $Y^{\rm CA}, Y^{\rm IA}$, and $Y^{\rm CDA}$ are the predictions of centralized, individual, and collaborative data analyses, respectively.
\par
All the numerical experiments are performed using MATLAB2019b.
\subsection{Experiment I: Artificial data}
\begin{figure}[!t]
\begin{tabular}{cc}
        \begin{minipage}[t]{0.45\hsize}
                \centering
                \includegraphics[scale=0.35, bb = 99 255 496 583]{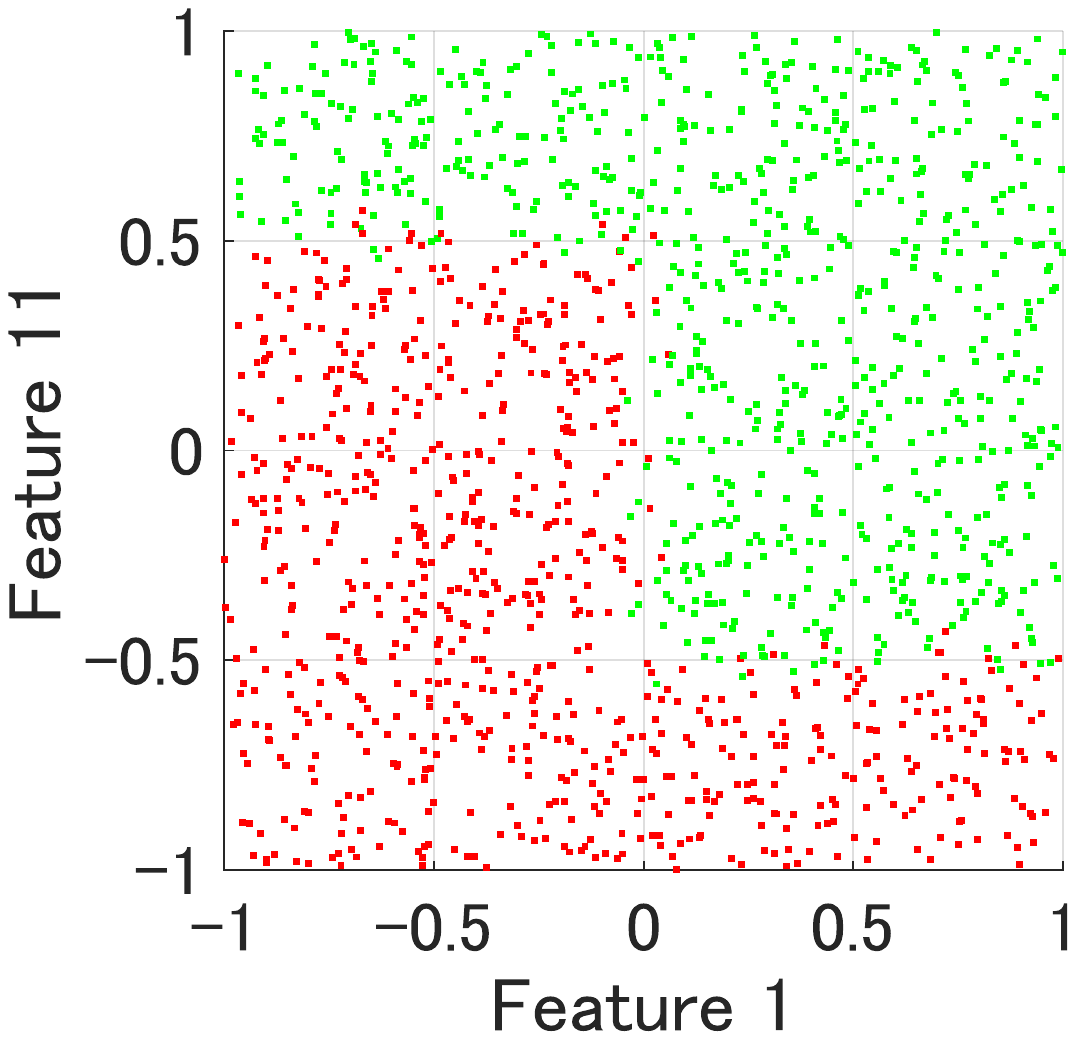}\\
                (a) All training dataset
        \end{minipage} &
        \begin{minipage}[t]{0.45\hsize}
                \centering
                \includegraphics[scale=0.35, bb = 99 255 496 583]{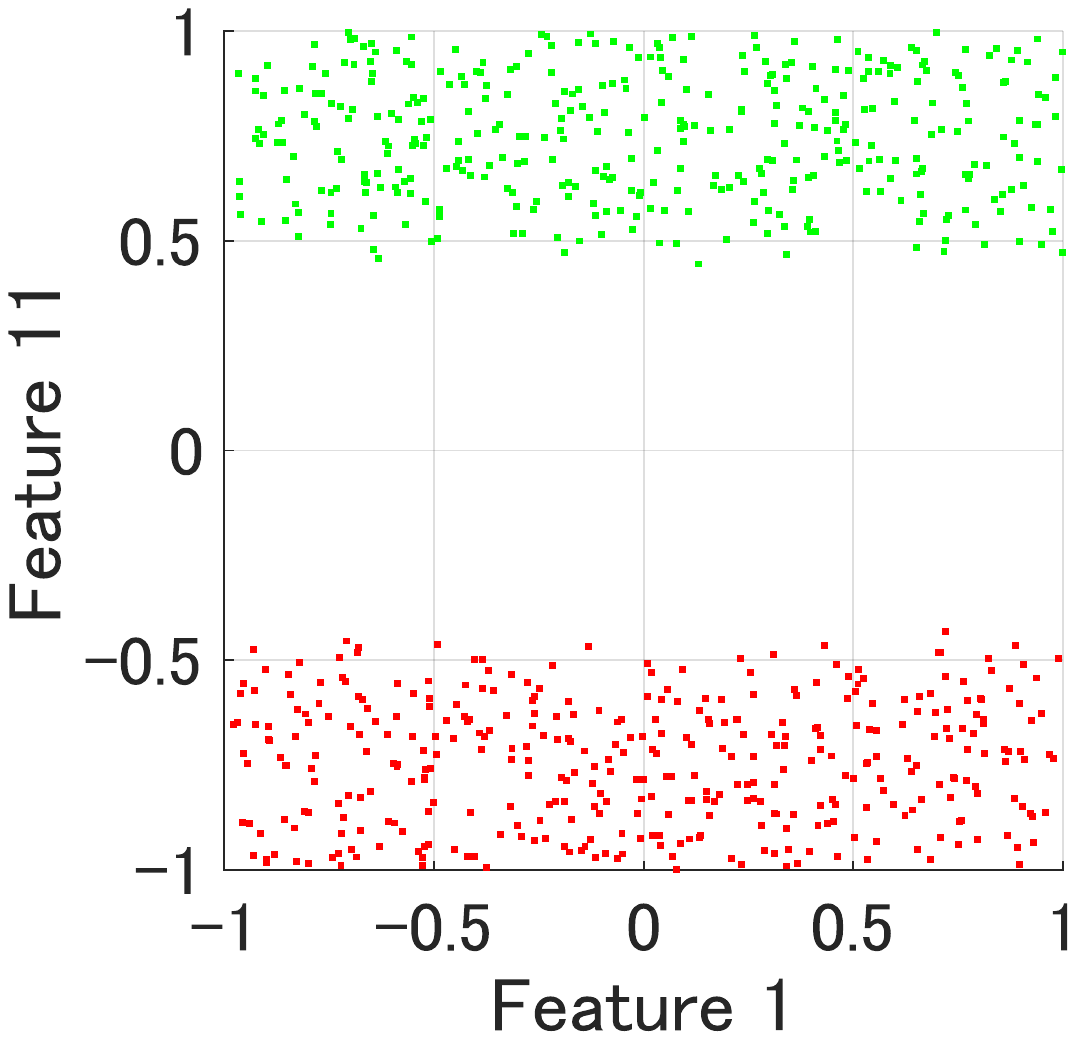}\\
                (b) Training dataset in the 1st group
        \end{minipage} \\
        \begin{minipage}[t]{0.45\hsize}
                \centering
                \includegraphics[scale=0.35, bb = 99 255 496 583]{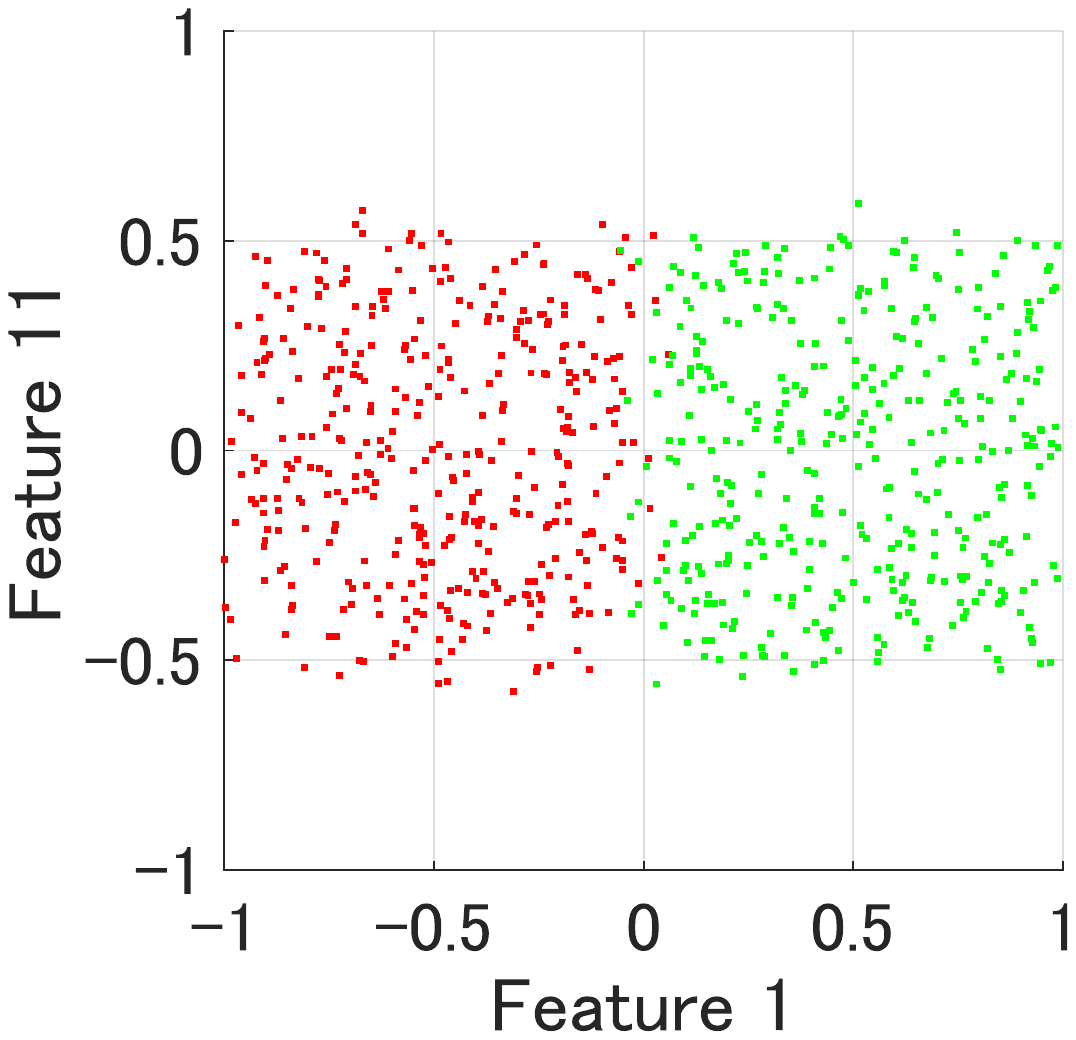}\\
                (c) Training dataset in the 2nd group
        \end{minipage} &
        \begin{minipage}[t]{0.45\hsize}
                \centering
                \includegraphics[scale=0.35, bb = 99 255 496 583]{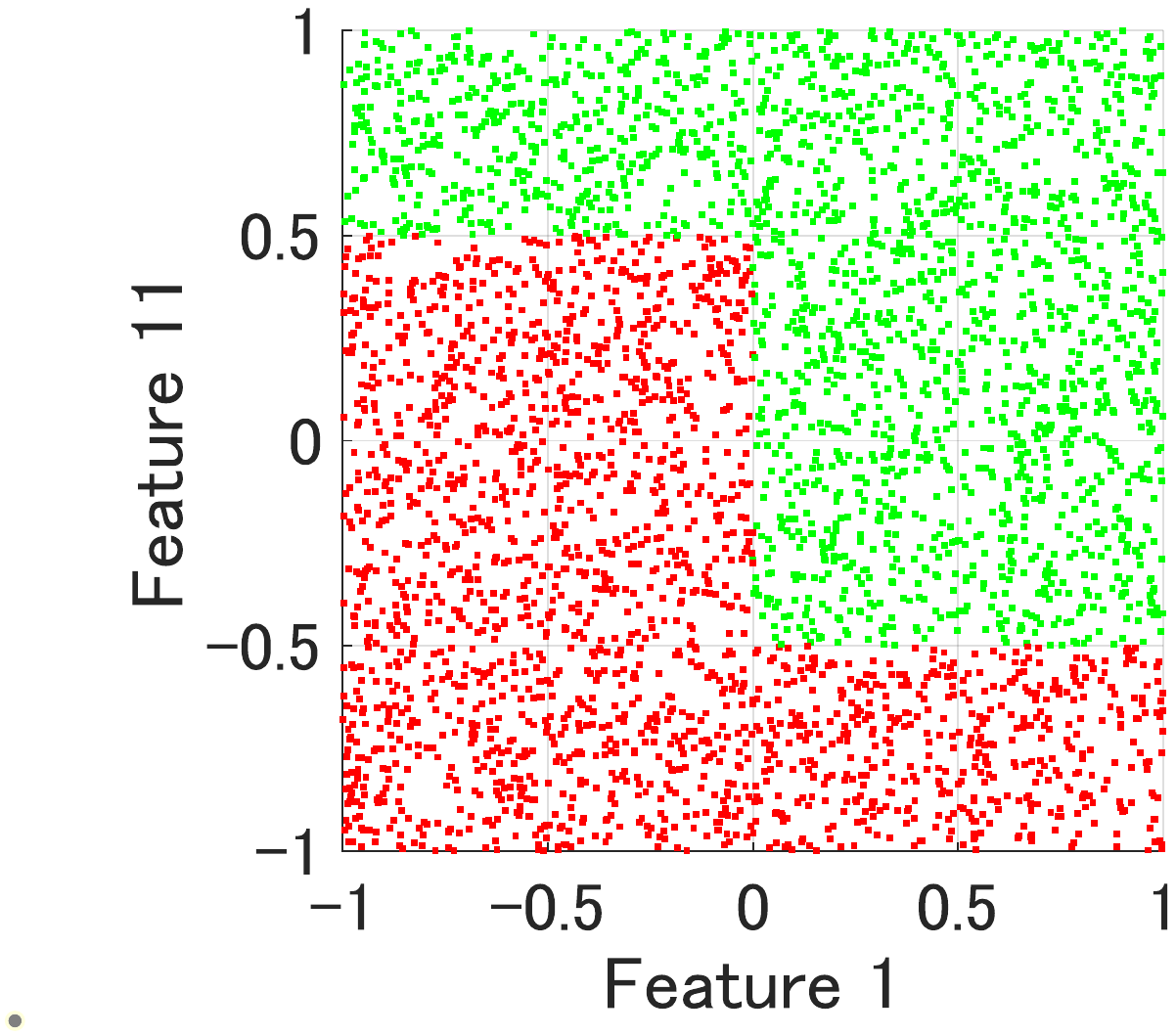}\\
                (d) Test dataset and its ground truth
        \end{minipage}
\end{tabular}
\caption{Features 1 and 2 of the training and test datasets for the artificial problem.}
\label{fig:train}
\end{figure}
We used a 20-dimensional artificial data for two-class classification.
Fig.~\ref{fig:train}(a) depicts features 1 and 11 of all the training datasets, where the number of samples is $n = 1,600$.
The other 18 dimensions have random values.
Note that only features 1 and 11 are necessary for classification.
\begin{figure}[!t]
\begin{tabular}{cc}
        \begin{minipage}[t]{0.45\hsize}
                \centering
                \includegraphics[scale=0.35, bb = 99 255 496 583]{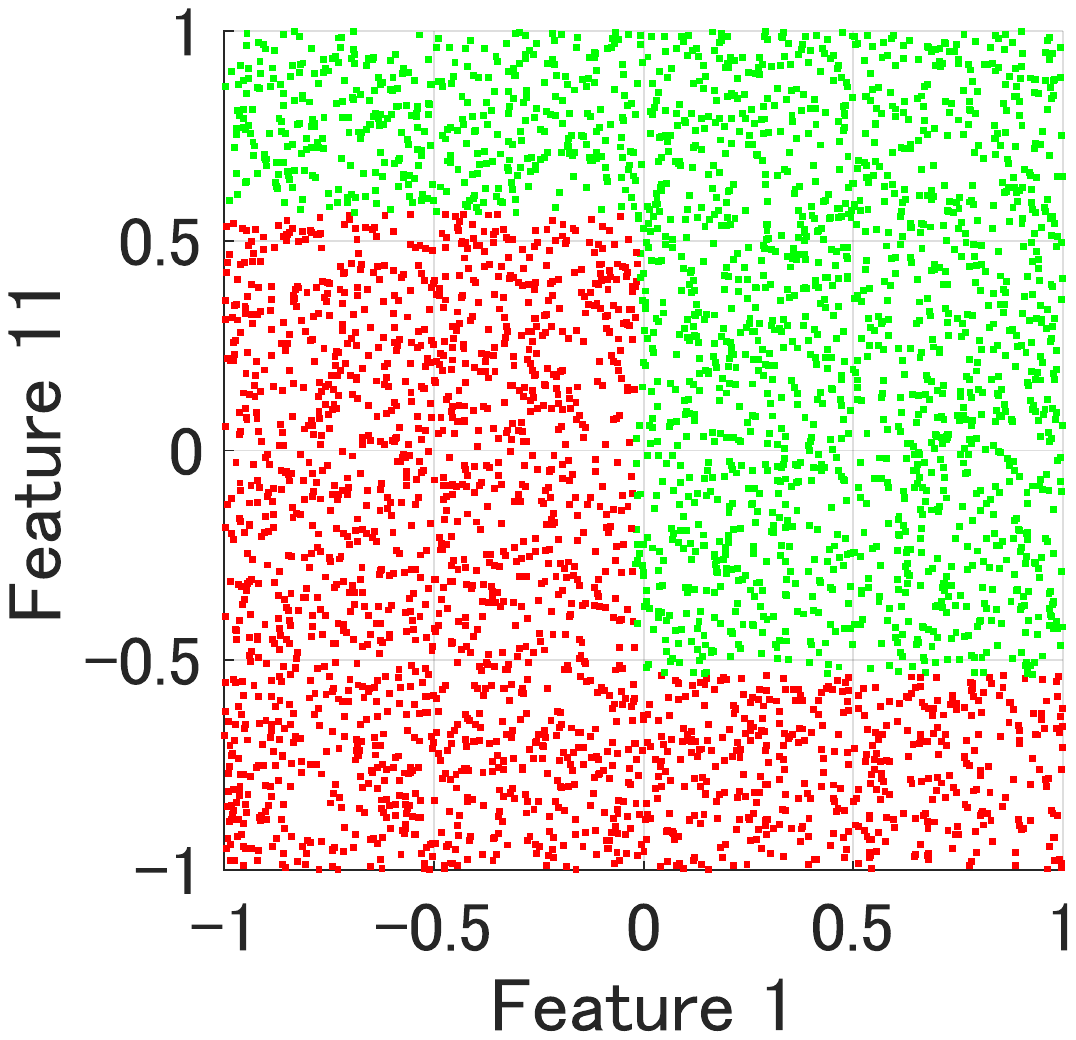}\\
                (a) Collaborative data analysis 
        \end{minipage} &
        \begin{minipage}[t]{0.45\hsize}
                \centering
                \includegraphics[scale=0.35, bb = 99 255 496 583]{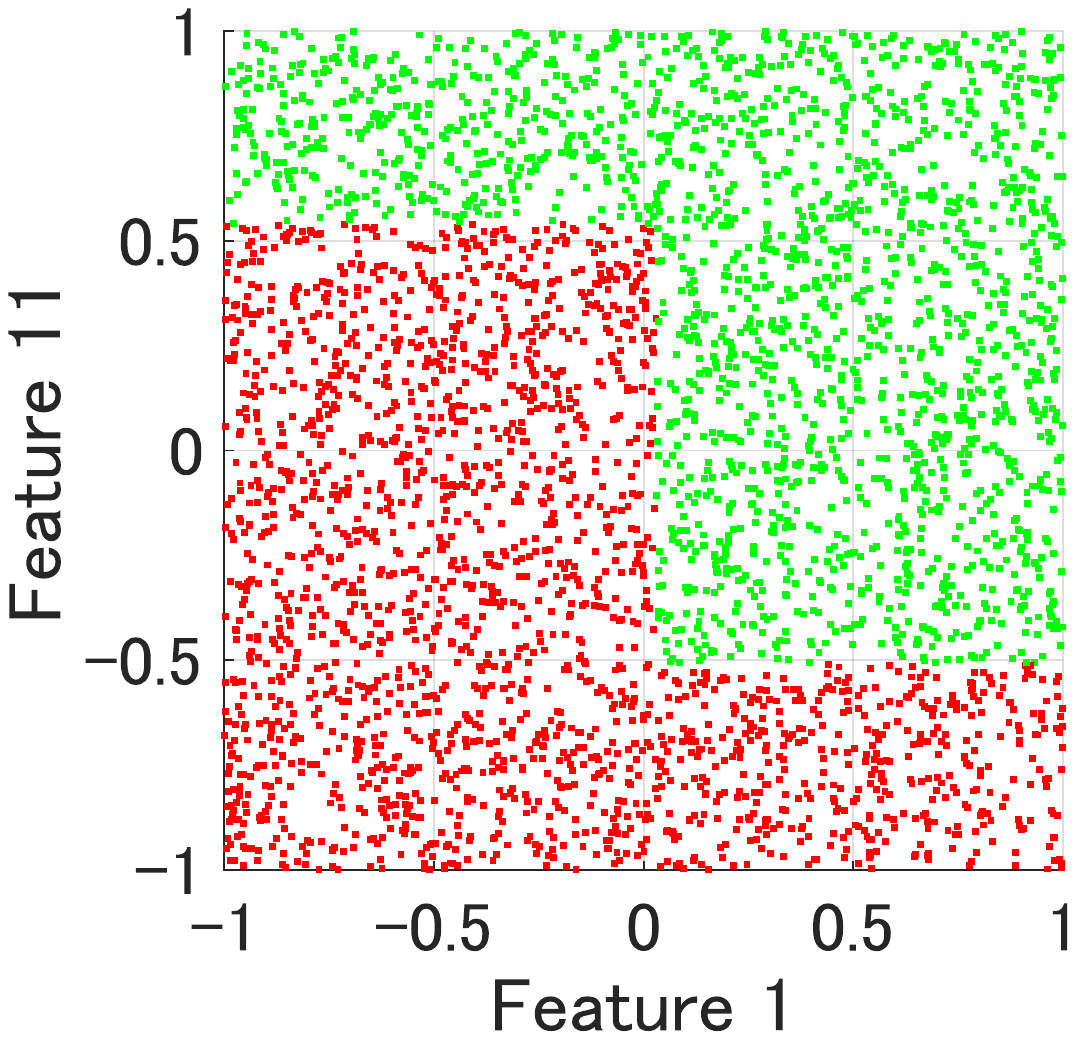}\\
                (b) Centralized analysis \\
        \end{minipage} \\
        \begin{minipage}[t]{0.45\hsize}
                \centering
                \includegraphics[scale=0.35, bb = 99 255 496 583]{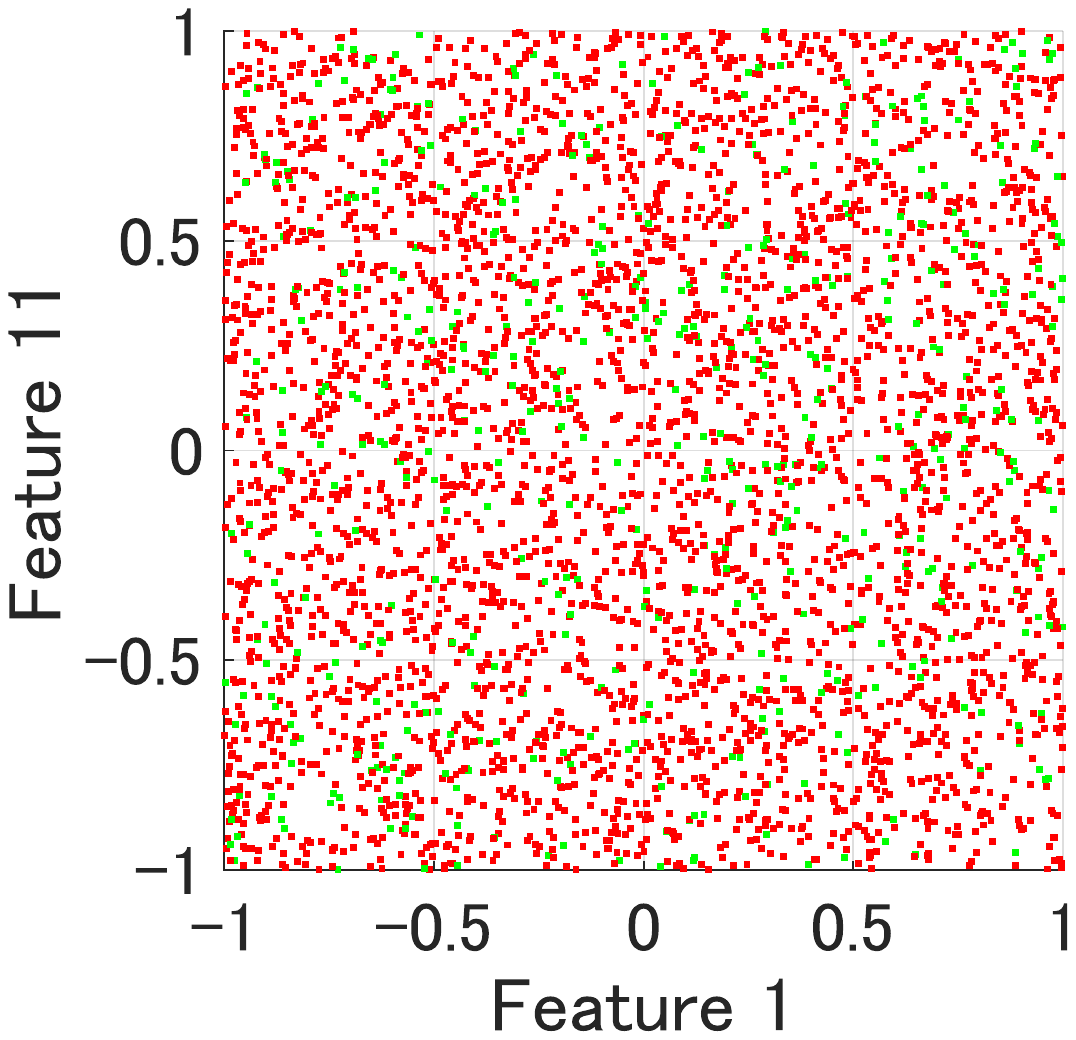}\\
                (c) Individual analysis for $X_{1,1}$ \\
        \end{minipage} &
        \begin{minipage}[t]{0.45\hsize}
                \centering
                \includegraphics[scale=0.35, bb = 99 255 496 583]{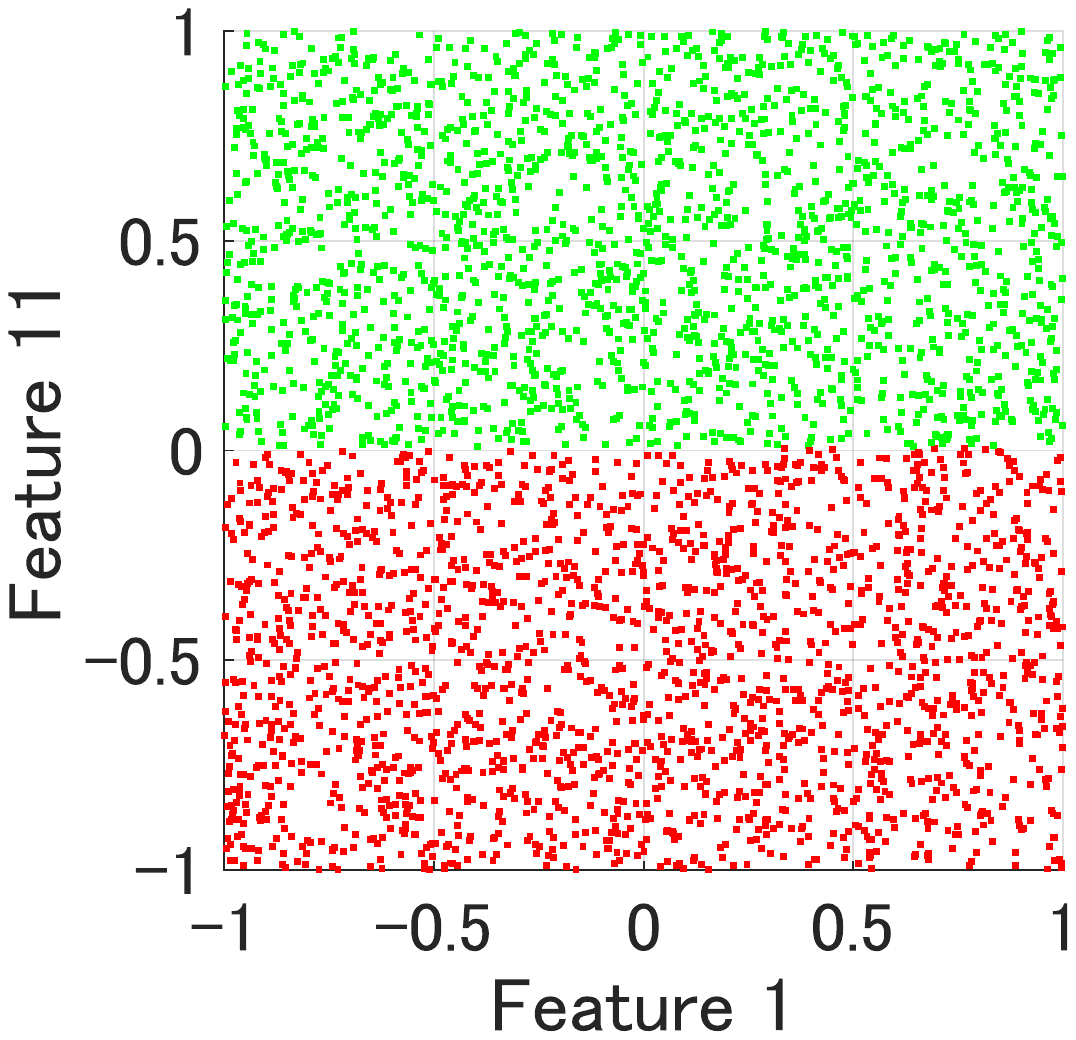}\\
                (d) Individual analysis for $X_{1,2}$ \\
        \end{minipage} \\
        \begin{minipage}[t]{0.45\hsize}
                \centering
                \includegraphics[scale=0.35, bb = 99 255 496 583]{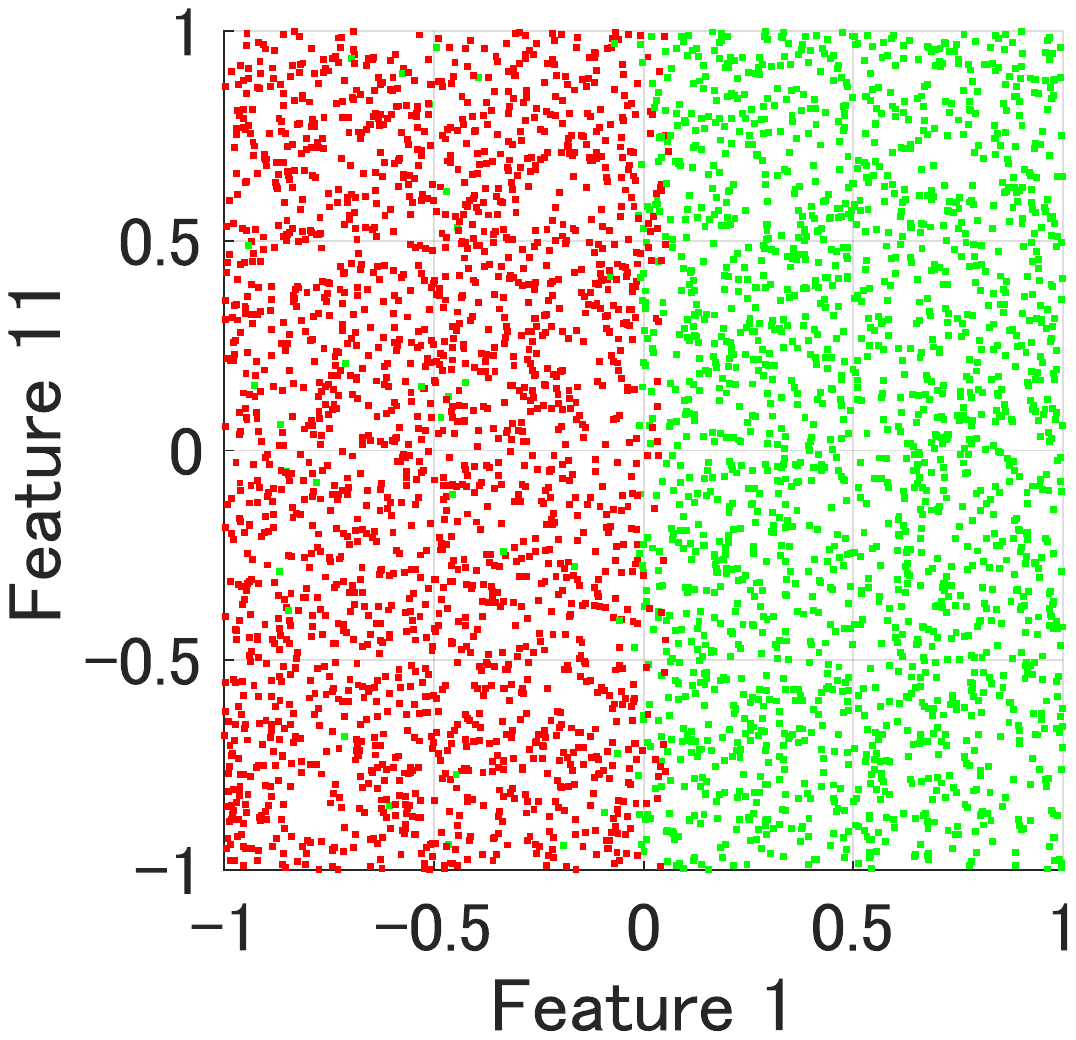}\\
                (e) Individual analysis for $X_{2,1}$ 
        \end{minipage} &
        \begin{minipage}[t]{0.45\hsize}
                \centering
                \includegraphics[scale=0.35, bb = 99 255 496 583]{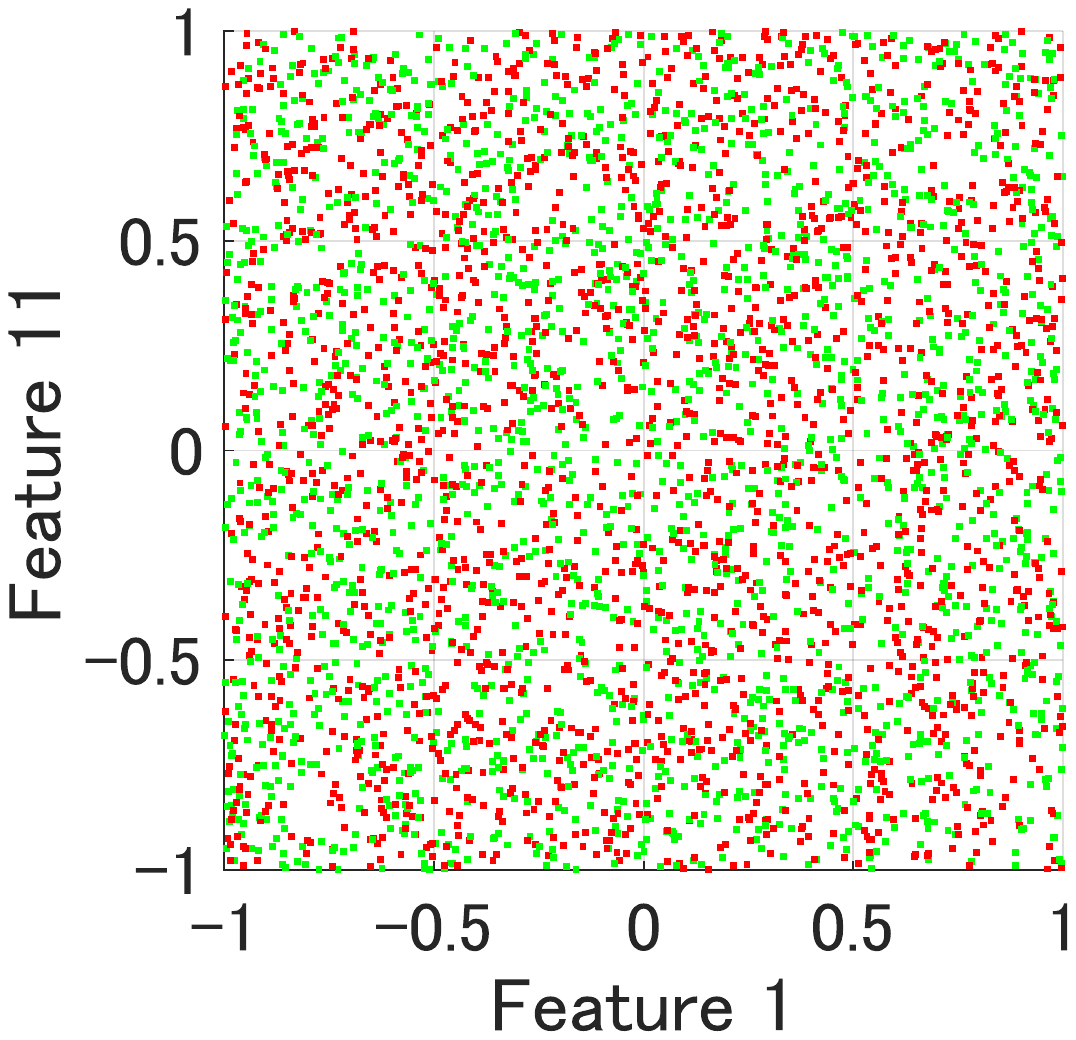}\\
                (f) Individual analysis for $X_{2,2}$ 
        \end{minipage}
\end{tabular}
\caption{Recognition results of centralized, individual, and collaborative data analyses for the artificial problem.}
\label{fig:result}
\end{figure}
\begin{table}[!t]
  \footnotesize
\caption{Recognition performance (average $\pm$ standard error) for the artificial problem.}
\label{table:ex1}
\begin{center}
\begin{tabular}{lcccccc}
\toprule
\multicolumn{1}{c}{Method} & & \multicolumn{1}{c}{NMI} & & \multicolumn{1}{c}{ACC} & & \multicolumn{1}{c}{Fidelity to CA}  \\ \cmidrule{1-1} \cmidrule{3-3} \cmidrule{5-5} \cmidrule{7-7}
Centralized  analysis & & $0.89 \pm 0.01$ & &  $98.37 \pm 0.14$ & &  $--$ \\
Individual   analysis & & $0.09 \pm 0.01$ & &  $62.14 \pm 1.95$ & &  $0.09 \pm 0.01$ \\
Collab. data analysis & & $0.83 \pm 0.02$ & &  $97.19 \pm 0.34$ & &  $0.82 \pm 0.01$ \\
\bottomrule
\end{tabular}
\end{center}
\end{table}
\par
We considered the case where the dataset in Fig.~\ref{fig:train}(a) is distributed into four parties: $c=d=2$ as
\begin{equation*}
  X = \left[ \begin{array}{cc}
      X_{1,1} & X_{1,2} \\
      X_{2,1} & X_{2,2} 
    \end{array}
  \right] \in \mathbb{R}^{1600 \times 20}, \quad
  X_{1,1}, X_{1,2}, X_{2,1}, X_{2,2} \in \mathbb{R}^{800 \times 10}.
\end{equation*}
For horizontal partitioning, the 1st group of parties $X_{1,1}, X_{1,2}$ is the dataset shown in Figs.~\ref{fig:train}(b) and the 2nd group of parties $X_{2,1}, X_{2,2}$ is the dataset shown in Fig.~\ref{fig:train}(c).
For the vertical partitioning, $X_{1,1}, X_{2,1}$ have the features 1--10 and $X_{1,2}, X_{2,2}$ have features 11--20.
Fig.~\ref{fig:train}(d) illustrates features 1 and 11 of the test dataset and their ground truth.
For the proposed method, we set the dimensionality of intermediate representations to $\widetilde{m}_{i,j} = 4$ for all parties.
\par
Fig.~\ref{fig:result} presents the recognition results and Table~\ref{table:ex1} shows the average and standard error of NMI, ACC, Fidelity to CA calculated across 10 trials.
From these results, we can observe that individual analysis does not obtain good recognition results.
This is because of the following reasons.
Since $X_{1,1}$ has feature~1 of the samples shown in Fig.~\ref{fig:train}(b) and $X_{2,2}$ has feature~11 of the samples shown in Fig.~\ref{fig:train}(c), the distributions of the two classes are overlapped.
Therefore, using only $X_{1,1}$ or $X_{2,2}$ cannot separate the two classes.
Moreover, $X_{1,2}$ has feature 11 of the samples shown in Fig.~\ref{fig:train}(b) and $X_{2,1}$ has feature 1 of the samples shown in Fig.~\ref{fig:train}(c).
Therefore, the classification boundaries for $X_{1,2}$ and $X_{2,1}$ are horizontal and vertical, respectively.
\par
On the other hand, when compared with individual analysis, the proposed collaborative data analysis (Fig.~\ref{fig:result}(a)) achieves good recognition results, which are comparable to the results of centralized analysis (Fig.~\ref{fig:result}(b)).
\subsection{Experiment II: Financial data}
\begin{figure}[!t]
\begin{center}
\begin{tabular}{cc}
        \begin{minipage}[t]{0.45\hsize}
                \centering
                \includegraphics[scale=0.25]{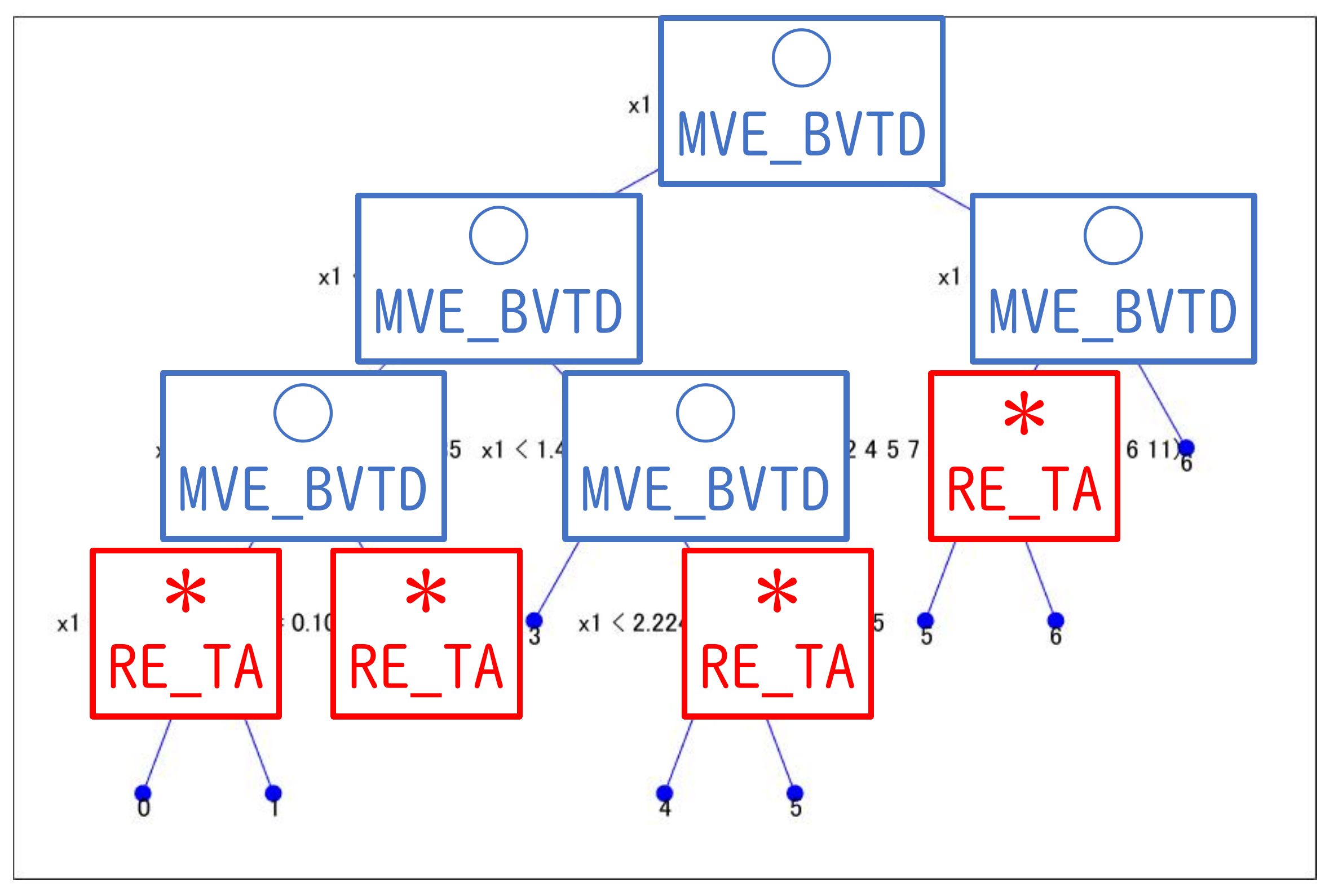}\\
                (a) Collaborative data analysis 
        \end{minipage} &
        \begin{minipage}[t]{0.45\hsize}
                \centering
                \includegraphics[scale=0.25]{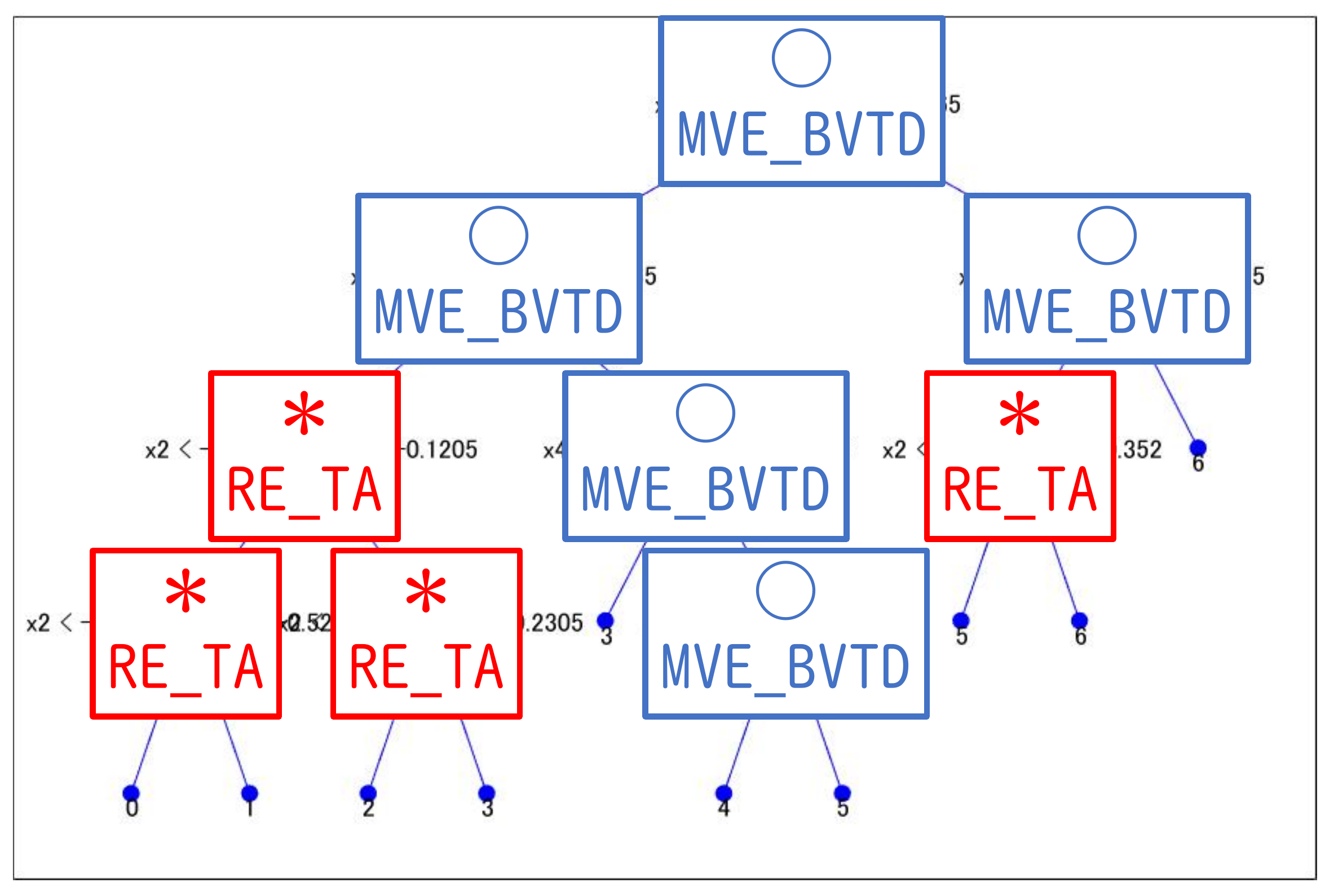}\\
                (b) Centralized analysis \\
        \end{minipage} \\
        \begin{minipage}[t]{0.45\hsize}
                \centering
                \includegraphics[scale=0.25]{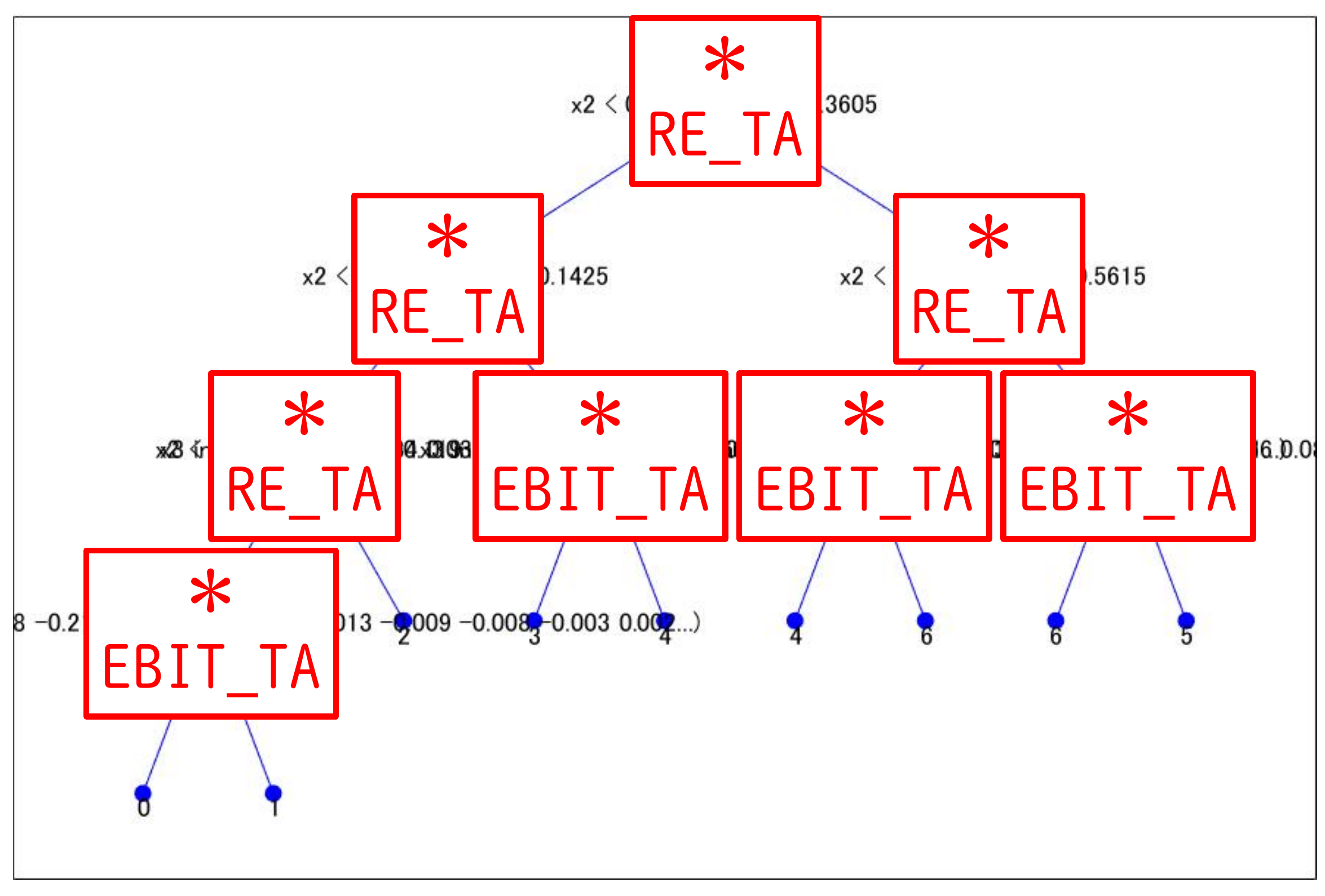}\\
                (c) Individual analysis for $X_{1,1}$ \\
        \end{minipage} &
        \begin{minipage}[t]{0.45\hsize}
                \centering
                \includegraphics[scale=0.25]{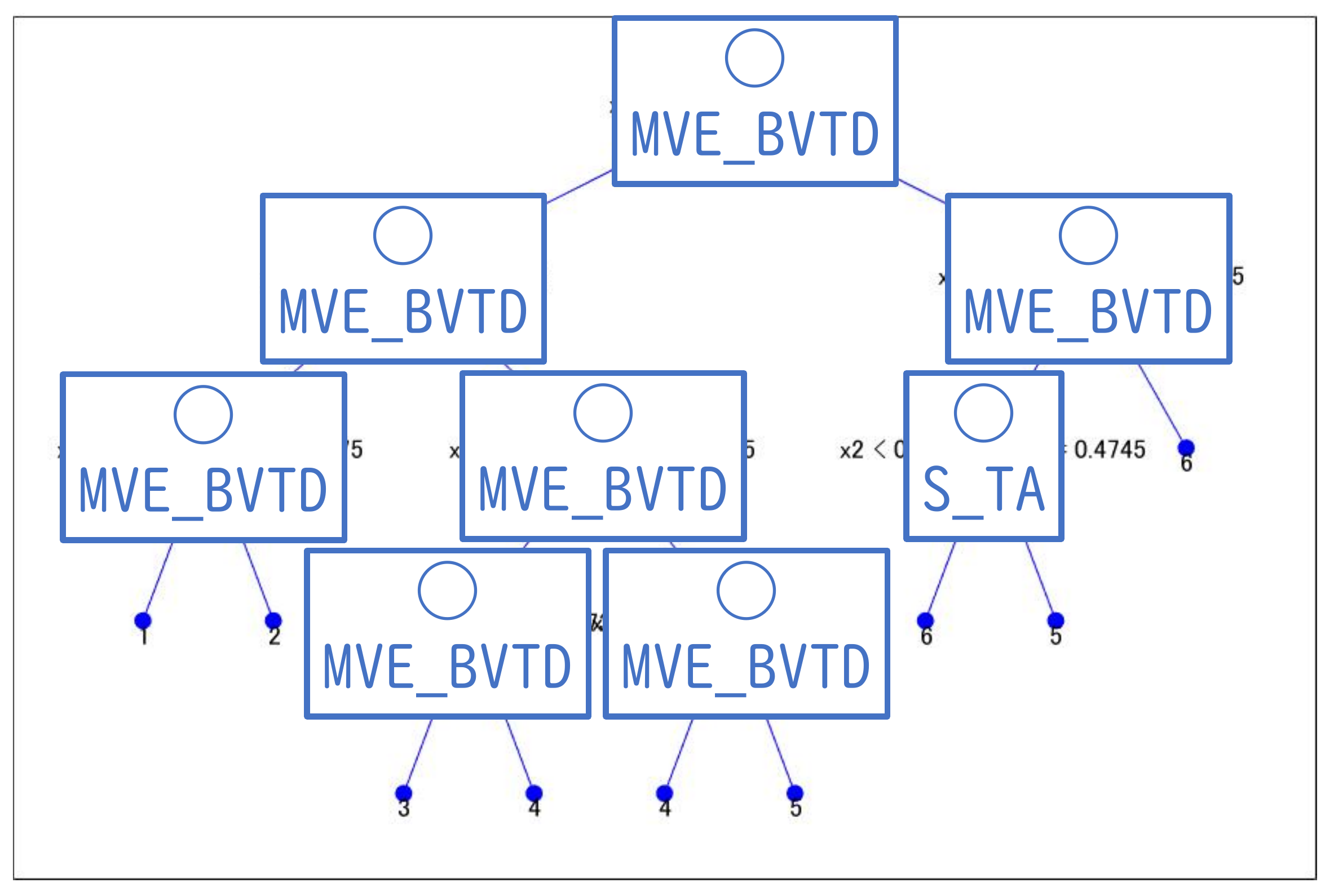}\\
                (d) Individual analysis for $X_{1,2}$ \\
        \end{minipage} \\
        \begin{minipage}[t]{0.45\hsize}
                \centering
                \includegraphics[scale=0.25]{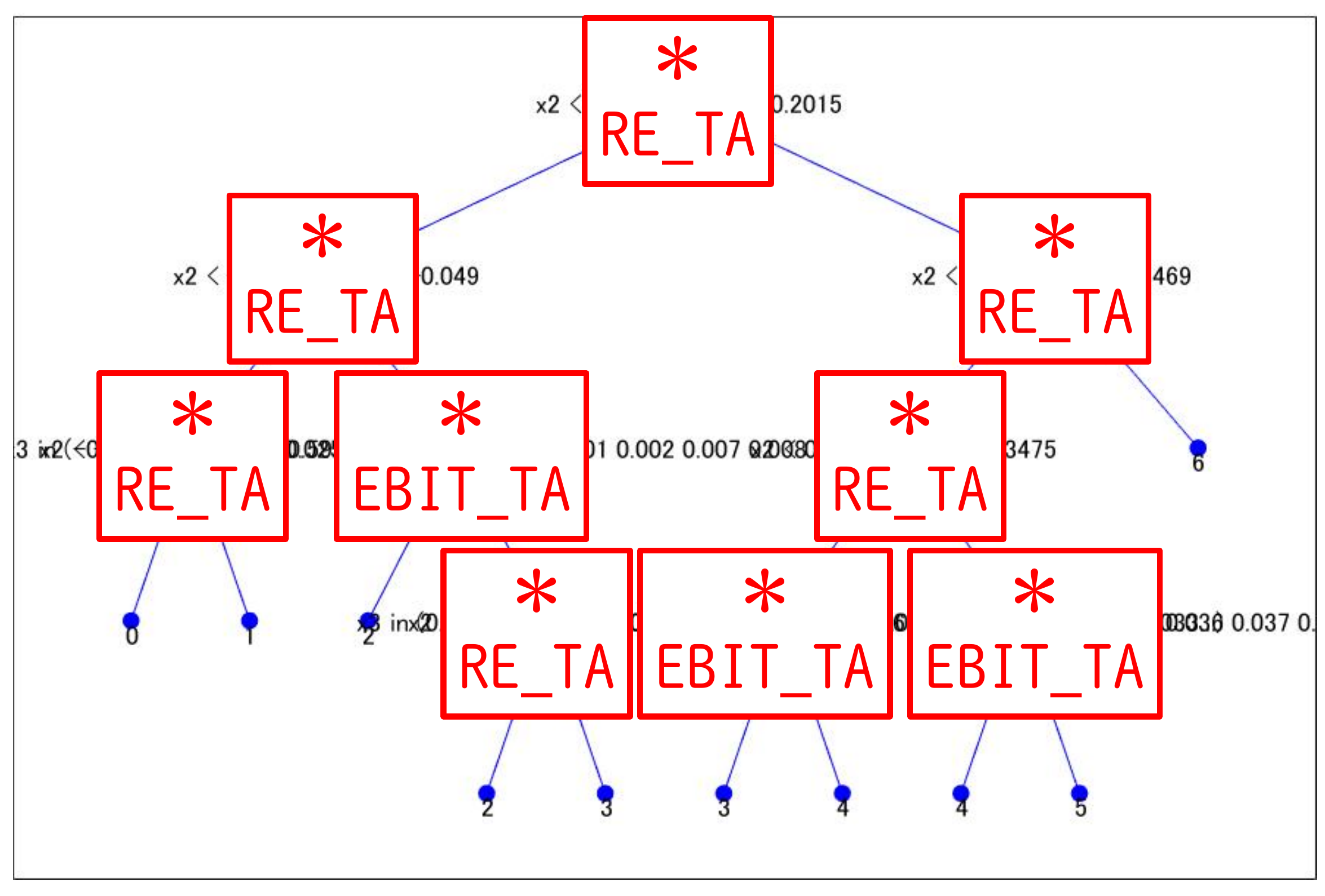}\\
                (e) Individual analysis for $X_{2,1}$ 
        \end{minipage} &
        \begin{minipage}[t]{0.45\hsize}
                \centering
                \includegraphics[scale=0.25]{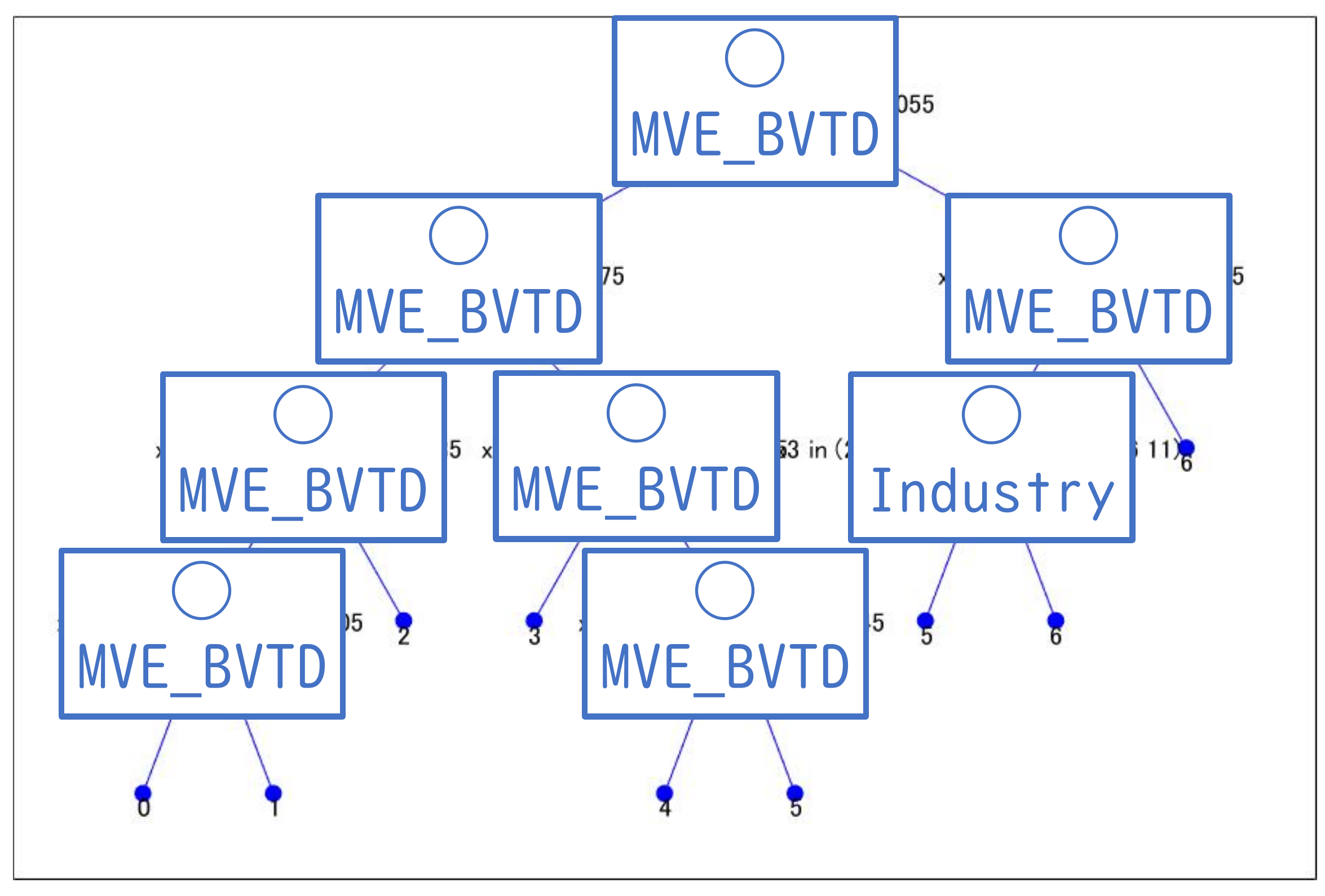}\\
                (f) Individual analysis for $X_{2,2}$ 
        \end{minipage} 
\end{tabular}
\caption{Decision tree of centralized, individual, and collaborative data analyses for the financial problem.}
\label{fig:tree}
\end{center}
\end{figure}
We used a credit rating dataset ``CreditRating\_Historical.dat'' from the MATLAB Statistics and Machine Learning Toolbox.
The dataset contains five financial ratios: Working capital / Total Assets (WC\_TA), Retained Earnings / Total Assets (RE\_TA), Earnings Before Interests and Taxes / Total Assets (EBIT\_TA), Market Value of Equity / Book Value of Total Debt (MVE\_BVTD), Sales / Total Assets (S\_TA), and industry sector labels from 1 to 12 for 3932 customers.
The dataset also includes credit ratings from ``AAA'' to ``CCC'' for all customers.
Note that this dataset is simulated and not real.
\begin{table}[!t]
  \footnotesize
\caption{Recognition performance (average $\pm$ standard error) for the financial problem.}
\label{table:ex2}
\begin{center}
\begin{tabular}{lcccccc}
\toprule
\multicolumn{1}{c}{Method} & & \multicolumn{1}{c}{NMI} & & \multicolumn{1}{c}{ACC} & & \multicolumn{1}{c}{Fidelity to CA}  \\ \cmidrule{1-1} \cmidrule{3-3} \cmidrule{5-5} \cmidrule{7-7}
Centralized  analysis & &  $0.59 \pm 0.00$ & & $69.59 \pm 0.51$ & & $--$ \\
Individual   analysis & &  $0.46 \pm 0.02$ & & $57.54 \pm 1.68$ & & $0.56 \pm 0.03$ \\
Collab. data analysis & &  $0.54 \pm 0.01$ & & $60.94 \pm 1.67$ & & $0.61 \pm 0.02$ \\
\bottomrule
\end{tabular}
\end{center}
\end{table}
\par
We aim to predict the credit rating using the five financial ratios and industry sector labels.
We considered the case where the training dataset with 3,000 samples is distributed into four parties: $c=d=2$ as
\begin{equation*}
  X = \left[ \begin{array}{cc}
      X_{1,1} & X_{1,2} \\
      X_{2,1} & X_{2,2} 
    \end{array}
  \right] \in \mathbb{R}^{3000 \times 6}, \quad
  X_{1,1}, X_{1,2}, X_{2,1}, X_{2,2} \in \mathbb{R}^{1500 \times 3},
\end{equation*}
where, $X_{1,1}, X_{2,1}$ have the 1st group of features WC\_TA, RE\_TA, and EBIT\_TA and $X_{1,2}, X_{2,2}$ have the 2nd group of features MVE\_BVTD, S\_TA, and Industry sector label as features.
\par
The obtained decision trees for centralized, individual, and collaborative data analyses are shown in Fig.~\ref{fig:tree}, while the average and standard error of NMI, ACC and Fidelity to CA across 10 trials are shown in Table~\ref{table:ex2}.
In Fig.~\ref{fig:tree}, the features with {\color{red} $\ast$} are in $X_{1,1}, X_{2,1}$ and the features with {\color{blue} $\circ$} are in $X_{1,2}, X_{2,2}$.
As shown in Fig.~\ref{fig:tree}, the proposed collaborative analysis (Fig.~\ref{fig:tree}(a)) has a tree with the same two features as centralized analysis, which belongs to different groups.
This cannot be achieved in individual analysis as shown in Fig.~\ref{fig:tree}(c)--(f).
\subsection{Experiment III: Real-world data}
\begin{table}[!p]
  \footnotesize
  \caption{Recognition performance (average $\pm$ standard error) for real-world problems.}
\label{table:result}
\begin{center}
\begin{tabular}{lclcccccc} 
\toprule
\multicolumn{1}{c}{Dataset} &  & \multicolumn{1}{c}{Method} &  & NMI &  & ACC &  & Fidelity to CA \\ \cmidrule{1-1} \cmidrule{1-1} \cmidrule{3-3} \cmidrule{5-5} \cmidrule{7-7} \cmidrule{9-9}
Carcinom     &  & CA  &  & $0.66\pm0.03$ & & $54.58\pm3.06$ & & $--$  \\
$\;\;m=9182$ &  & IA  &  & $0.50\pm0.04$ & & $40.67\pm3.12$ & & $0.48\pm0.04$  \\
$\;\;n\;=174$&  & CDA &  & $0.66\pm0.04$ & & $54.06\pm5.30$ & & $0.73\pm0.04$  \\ \cmidrule{1-1} \cmidrule{3-3} \cmidrule{5-5} \cmidrule{7-7} \cmidrule{9-9}
CLL-SUB-111  &  & CA  &  & $0.22\pm0.03$ & & $60.40\pm2.86$ & & $--$  \\
$\;\;m=11340$&  & IA  &  & $0.16\pm0.02$ & & $56.81\pm1.64$ & & $0.14\pm0.02$  \\
$\;\;n\;=111$&  & CDA &  & $0.29\pm0.08$ & & $52.06\pm5.43$ & & $0.18\pm0.06$  \\ \cmidrule{1-1} \cmidrule{3-3} \cmidrule{5-5} \cmidrule{7-7} \cmidrule{9-9}
GLA-BRA-180  &  & CA  &  & $0.37\pm0.05$ & & $62.22\pm3.74$ & & $--$  \\
$\;\;m=49151$&  & IA  &  & $0.28\pm0.02$ & & $55.74\pm1.75$ & & $0.30\pm0.02$  \\
$\;\;n\;=180$&  & CDA &  & $0.33\pm0.04$ & & $61.67\pm2.41$ & & $0.37\pm0.03$  \\ \cmidrule{1-1} \cmidrule{3-3} \cmidrule{5-5} \cmidrule{7-7} \cmidrule{9-9}
jaffe        &  & CA  &  & $0.68\pm0.02$ & & $38.95\pm1.43$ & & $--$  \\
$\;\;m=676$  &  & IA  &  & $0.64\pm0.01$ & & $42.36\pm1.51$ & & $0.55\pm0.03$  \\
$\;\;n\;=213$&  & CDA &  & $0.59\pm0.04$ & & $31.46\pm2.46$ & & $0.64\pm0.04$  \\ \cmidrule{1-1} \cmidrule{3-3} \cmidrule{5-5} \cmidrule{7-7} \cmidrule{9-9}
leukemia     &  & CA  &  & $0.74\pm0.14$ & & $94.64\pm2.95$ & & $--$  \\
$\;\;m=7129$ &  & IA  &  & $0.39\pm0.06$ & & $80.39\pm2.39$ & & $0.36\pm0.06$  \\
$\;\;n\;=72$ &  & CDA &  & $0.39\pm0.10$ & & $81.61\pm3.94$ & & $0.38\pm0.04$  \\ \cmidrule{1-1} \cmidrule{3-3} \cmidrule{5-5} \cmidrule{7-7} \cmidrule{9-9}
lung         &  & CA  &  & $0.69\pm0.05$ & & $88.14\pm2.08$ & & $--$  \\
$\;\;m=3312$ &  & IA  &  & $0.52\pm0.03$ & & $78.06\pm1.50$ & & $0.48\pm0.03$  \\
$\;\;n\;=203$&  & CDA &  & $0.64\pm0.05$ & & $86.74\pm1.57$ & & $0.57\pm0.08$  \\ \cmidrule{1-1} \cmidrule{3-3} \cmidrule{5-5} \cmidrule{7-7} \cmidrule{9-9}
pixraw10P    &  & CA  &  & $0.68\pm0.02$ & & $38.00\pm1.79$ & & $--$  \\
$\;\;m=10000$&  & IA  &  & $0.63\pm0.04$ & & $41.00\pm3.12$ & & $0.46\pm0.04$  \\
$\;\;n=100$  &  & CDA &  & $0.61\pm0.03$ & & $29.00\pm0.89$ & & $0.62\pm0.06$  \\ \cmidrule{1-1} \cmidrule{3-3} \cmidrule{5-5} \cmidrule{7-7} \cmidrule{9-9}
Prostate\_GE &  & CA  &  & $0.40\pm0.10$ & & $84.09\pm3.63$ & & $--$  \\
$\;\;m=5966$ &  & IA  &  & $0.28\pm0.04$ & & $75.30\pm2.20$ & & $0.26\pm0.03$  \\
$\;\;n\;=102$&  & CDA &  & $0.31\pm0.07$ & & $78.36\pm2.37$ & & $0.34\pm0.06$  \\ \cmidrule{1-1} \cmidrule{3-3} \cmidrule{5-5} \cmidrule{7-7} \cmidrule{9-9}
TOX-171      &  & CA  &  & $0.37\pm0.03$ & & $59.70\pm2.62$ & & $--$  \\
$\;\;m=5789$ &  & IA  &  & $0.30\pm0.01$ & & $49.52\pm1.51$ & & $0.29\pm0.02$  \\
$\;\;n\;=171$&  & CDA &  & $0.34\pm0.04$ & & $49.68\pm3.72$ & & $0.36\pm0.05$  \\ \cmidrule{1-1} \cmidrule{3-3} \cmidrule{5-5} \cmidrule{7-7} \cmidrule{9-9}
warpAR10P    &  & CA  &  & $0.64\pm0.02$ & & $30.00\pm2.96$ & & $--$  \\
$\;\;m=2400$ &  & IA  &  & $0.59\pm0.02$ & & $34.87\pm1.39$ & & $0.52\pm0.02$  \\
$\;\;m=130$  &  & CDA &  & $0.59\pm0.03$ & & $30.77\pm3.61$ & & $0.53\pm0.05$  \\ \cmidrule{1-1} \cmidrule{3-3} \cmidrule{5-5} \cmidrule{7-7} \cmidrule{9-9}
\bottomrule
\end{tabular}
\end{center}
\end{table}

We next evaluate the performances of centralized, individual, and collaborative data analyses on the binary and multi-class classification problems obtained from \cite{lecun1998mnist,samaria2994parameterisation} and feature selection datasets \footnote{available at \url{http://featureselection.asu.edu/datasets.php.}}.
\par
We considered the case where the dataset is distributed into six parties: $c=2$ and $d=3$.
The performance of each method is evaluated by using a five-fold cross-validation framework.
For the proposed method, we set $\widetilde{m}_{i,j} = 15$.
\par
The numerical results of the centralized analysis, an average of the individual analysis and the proposed method for 10 test problems are presented in Table~\ref{table:result}.
We can observe from Table~\ref{table:result} that recognition performance of the proposed method is better than that of individual analysis and comparable to that of centralized analysis on most datasets.
\section{Conclusions}
\label{sec:conclusion}
To address the needs of distributed data analysis and achieve interpretability, we proposed an interpretable non-model sharing collaborative data analysis on distributed data.
The proposed method generated an interpretable model for distributed data by sharing intermediate representations without revealing the private data and the model.
The obtained interpretable model was based on the whole features of distributed data, which cannot be achieved in individual analysis.
Numerical experiments on both artificial and real-world data showed that the proposed method constructed an interpretable model with better recognition performance than individual analysis and comparable to centralized analysis.
\par
The distributed data analysis and the interpretable model construction are essential and important challenges in real-world situations including medial, financial, and manufacturing data analyses.
The proposed interpretable collaborative data analysis would be a breakthrough technology for such kinds of distributed data analysis.
\par
In our future studies, we will further analyze the privacy and confidentiality concerns and the accuracy of the proposed method.
Moreover, practical techniques for improving the performance of the proposed method including other suitable anchor data will be investigated.
The authors will also apply the proposed method to practical distributed data in other fields, such as medical or manufacturing, and evaluate its recognition performance.
\section*{Acknowledgements}
The  work was supported in part by the New Energy and Industrial Technology Development Organization (NEDO).
The work of the first author was supported in part by the Japan Science and Technology Agency (JST), ACT-I (No. JPMJPR16U6) and the Japan Society for the Promotion of Science (JSPS), Grants-in-Aid for Scientific Research (Nos. 17K12690, 19KK0255).
The work of the fourth author was supported in part by the Japan Society for the Promotion of Science (JSPS), Grants-in-Aid for Scientific Research (No. 18H03250).

\bibliography{mybibfile}
\bibliographystyle{elsart-num-sort}

\end{document}